\documentclass[onecolumn]{IEEEtran}
\usepackage{hyperref}
\IEEEoverridecommandlockouts
\usepackage[a4paper, margin=1in]{geometry} 
\usepackage{cite}
\usepackage{amsmath,amssymb,amsfonts}
\usepackage{algorithmic}
\usepackage{graphicx}
\usepackage{textcomp}
\usepackage{xcolor}
\def\BibTeX{{\rm B\kern-.05em{\sc i\kern-.025em b}\kern-.08em
    T\kern-.1667em\lower.7ex\hbox{E}\kern-.125emX}}

\title{\textbf{A Comparative Study of Explainable AI Methods: Model-Agnostic vs. Model-Specific Approaches}}

\author{\IEEEauthorblockN{Keerthi Devireddy} \\
\IEEEauthorblockA{
Email: keerthidevireddy0289\@gmail.com}
}

\begin{document}
\maketitle

\section{\textbf{Introduction to Explainable AI (XAI)}}
\subsection*{\textbf{Introduction}}
Deep learning models, especially convolutional neural networks (CNNs), have achieved remarkable performance across various domains, including image recognition, natural language processing, and medical diagnostics. However, their decision-making process remains a black-box, making it challenging to understand how and why certain predictions are made.
Explainable AI (XAI) addresses this issue by providing interpretability to complex models, helping users, researchers, and regulatory bodies build trust in AI-driven decisions. The need for interpretability is particularly crucial in:
\begin{itemize}
    \item \textbf{Healthcare:} AI-based diagnostics must provide explanations to assist doctors in decision-making.
    \item \textbf{Autonomous Systems:} Self-driving cars must justify their decisions for safety compliance.
    \item \textbf{Finance:} Loan approvals and fraud detection systems need explainability for accountability.
    \item \textbf{Legal \& Ethical AI:} Ensuring AI decisions are fair, unbiased, and compliant with regulations.
\end{itemize}
XAI techniques shed light on model decisions, helping identify biases, debugging models, and ensuring compliance with ethical AI guidelines.
\subsection*{\textbf{Why is XAI Important?}}
XAI is essential for the following reasons:
\begin{itemize}
    \item \textbf{Trust \& Transparency} – Users can trust AI-driven decisions when they are interpretable.
    \item \textbf{Debugging \& Model Improvement} – Helps identify errors or biases in deep learning models.
    \item \textbf{Regulatory Compliance} – Ensures models adhere to AI regulations like GDPR and AI Act.
    \item \textbf{Better Human-AI Collaboration} – Enables domain experts to make informed decisions based on model outputs.
\end{itemize}
\subsection*{\textbf{Mathematical Definition of Interpretability}}
The interpretability of a model can be defined as the function: 
\[\textbf{\textit{I(M)=f(M,D,C)}} \]
Where:
\begin{itemize}
    \item \(M\) is the model.
    \item \(D\) is the dataset.
    \item \(C\) represents the constraints (such as domain knowledge, computational efficiency).
    \item \(f\) measures the degree to which a human can understand and trust the model.
\end{itemize}
A well-designed XAI method should maximize I(M) without compromising accuracy.
\section{\textbf{Model-Agnostic vs. Model-Specific XAI}}
\subsection*{\textbf{Overview of Explainability Methods}}
Explainable AI (XAI) techniques can be categorized based on their dependence on the model structure. The two broad categories are:
\begin{itemize}
    \item \textbf{Model-Agnostic Methods} - 
    These methods work independently of the underlying model architecture, making them flexible and applicable across different machine learning models (e.g., decision trees, neural networks, etc.). However, they often require post hoc analysis, making them computationally expensive.
    \item \textbf{Model-Specific Methods} -
    These methods leverage the internal structure of a specific model (e.g., convolutional neural networks, decision trees) to provide more precise and architecture-dependent explanations.
\end{itemize}
\begin{table}[h]
    \centering
    \begin{tabular}{|p{4cm}|p{5cm}|p{5cm}|}  
        \hline
        \textbf{Feature} & \textbf{Model-Agnostic Methods} & \textbf{Model-Specific Methods} \\ \hline
        Applicability & Any ML model (CNNs, Decision Trees, etc.) & Only specific models (e.g., CNNs) \\ \hline
        Flexibility & High & Low \\ \hline
        Computational Cost & Higher due to post-hoc analysis & Lower, integrated into the model \\ \hline
        Interpretability & Provides broad explanations but can be less precise & More precise explanations for specific architectures \\ \hline
    \end{tabular}
    \label{tab:model_comparison} 
\end{table}
For this study, I have chosen two model-agnostic methods (LIME and SHAP) and two model-specific methods (Grad-CAM and Guided Backpropagation) to compare their effectiveness in explaining deep learning-based image classification.
\subsection*{\textbf{Selected XAI Methods}}
\subsection*{\textbf{Model-Agnostic Methods}}
Model-agnostic methods allow us to explain any black-box model by approximating it with a simpler interpretable model or analyzing feature contributions. They are particularly useful when we do not have direct access to the internals of a model or when working with proprietary systems. 
\subsection*{\textbf{1. Local Interpretable Model-Agnostic Explanations (LIME)}}
LIME explains individual predictions by training a simplified surrogate model that approximates the complex model’s behavior around a given instance. The key idea is to perturb the input data, observe changes in the model's predictions, and fit a local linear model to approximate decision boundaries. \\
\\
\textbf{Workflow of LIME:}
\begin{itemize}
    \item \textbf{Perturb the Input:} Generate multiple variations of the input image by modifying superpixels or pixel regions.
    \item \textbf{Predict with the Model:} Pass these perturbed images through the deep learning model and collect predictions.
    \item \textbf{Train a Surrogate Model:} Fit a simple linear model that mimics the black-box model’s behavior around the instance of interest.
    \item \textbf{Interpretation:} The coefficients of the surrogate model highlight the most important regions contributing to the prediction.
\end{itemize}
\textbf{Mathematical Formulation}\\
LIME optimizes the following function to ensure the surrogate model g is both faithful to the original model f and interpretable: \\
\[g(x) \approx f(x)\]
for a set of perturbed instances \( x' \) weighted by proximity \( \pi_x(x') \). The optimization objective for the surrogate model is:

\begin{equation}
\arg\min_{g \in G} \mathcal{L}(f, g, \pi_x) + \Omega(g)
\end{equation}
where:
\begin{itemize}
    \item \( \mathcal{L}(f, g, \pi_x) \) is the loss function ensuring \(g\) mimics \(f\) for perturbations around \(x\).
    \item \( \pi_x \) assigns higher weights to perturbed instances closer to \( x \).
    \item \( \Omega(g) \)  regularizes \(g \) to remain interpretable (e.g., sparsity constraint).
\end{itemize}
LIME creates a local interpretable model \(g \) that approximates the complex function \(f \) in the vicinity of a given input \(x \). The loss function \( \mathcal{L}(f, g, \pi_x) \) ensures that \(g \) closely resembles \(f \), while the regularization term  \( \Omega(g) \) enforces interpretability by keeping \(g \) simple (e.g., linear models or decision trees).
\vspace{5pt}  
\subsection*{\textbf{2. Shapley Additive Explanations (SHAP)}}
SHAP explains model predictions using concepts from cooperative game theory, attributing feature importance based on their contribution to different coalitions of features. Unlike LIME, which approximates local behavior, SHAP ensures global and local interpretability. \\
\\
\textbf{Workflow of SHAP:}
\begin{itemize}
    \item \textbf{Form Feature Coalitions:} Consider different subsets of features (e.g., pixels or regions of an image).
    \item \textbf{Measure Contribution:} Compute how adding or removing a feature affects model predictions.
    \item \textbf{Compute Shapley Values:} Assign a score to each feature based on its average marginal contribution across all subsets.
    \item \textbf{Visualization:} The output heatmap shows which regions positively or negatively influence the prediction.
\end{itemize}
\textbf{Mathematical Definition} \\
The Shapley value for a feature \(i\) is computed as: \\
\begin{equation}
\phi_i = \sum_{S \subseteq N \setminus \{i\}} \frac{|S|!(|N|-|S|-1)!}{|N|!} \left( f(S \cup \{i\}) - f(S) \right)
\end{equation}
where:
\begin{itemize}
    \item \(S\) is a subset of features,
    \item \(N\) is the set of all features,
    \item \(f(S)\) is the model’s output when only features in \(S\) are considered.
\end{itemize}
SHAP provides strong theoretical guarantees and is widely used to interpret deep learning predictions. SHAP values are derived from cooperative game theory and fairly distribute the contribution of each feature \(i\) by considering all possible subsets \(S\) of features. The term \(f(S \cup \{i\}) - f(S)\) represents the change in model output when feature \(i\) is included, weighted by a factor that ensures fair allocation across different subset sizes.

\subsection*{\textbf{Model-Specific XAI Methods}}
Model-specific methods take advantage of internal gradients and activations to provide more precise and architecture-dependent explanations. 
\subsection*{\textbf{1. Gradient-weighted Class Activation Mapping (Grad-CAM)}}
Grad-CAM visualizes the most influential regions of an image by computing gradients of the class score w.r.t. feature maps in the CNN. It helps in localizing discriminative features used by the network. \\
\\
\textbf{Workflow of Grad-CAM:}
\begin{itemize}
    \item \textbf{Forward Pass:} Compute feature maps from a convolutional layer.
    \item \textbf{Compute Gradients:} Calculate class-specific gradients w.r.t. these feature maps.
    \item \textbf{Weight Aggregation:} Compute the importance of each feature map using global average pooling.
    \item \textbf{Heatmap Generation:} Apply ReLU to generate a positive activation map overlay on the original image.
\end{itemize}

\textbf{Mathematical Formulation }
The Grad-CAM heatmap is computed as:
\begin{equation}
L^c = ReLU \left( \sum_k \alpha_k^c A^k \right)
\end{equation}
where: \\
\begin{itemize}
    \item \( A^k\) is the activation map for convolutional layer \(k\).
    \item \(\alpha_k^c\) represents the importance weights:
\end{itemize}
\begin{equation}
\alpha_k^c = \frac{1}{Z} \sum_i \sum_j \frac{\partial y^c}{\partial A^k_{ij}}
\end{equation}
\begin{itemize}
    \item \(Z\) is the number of pixels in the feature map.
\end{itemize}
Grad-CAM helps in visualizing class-discriminative regions for CNN-based models. 
Grad-CAM highlights image regions most relevant to a class prediction by computing the importance weights \(\alpha_k^c\), which are obtained via gradients of the class score \(y^c\) with respect to feature maps \(A^k\). The final heatmap is generated by applying ReLU to the weighted feature maps, ensuring only positive influences are visualized.
\subsection*{\textbf{Guided Backpropagation}}
Guided Backpropagation modifies standard backpropagation to suppress negative gradients, ensuring only positive, class-relevant features are propagated. \\
\\
\textbf{Workflow of Guided Backpropagation:}
\begin{itemize}
    \item \textbf{Compute Gradients:} Perform standard backpropagation through the network.
    \item \textbf{Suppress Negative Signals:} Restrict gradient flow by setting negative gradients to zero.
    \item \textbf{Generate Relevance Map:} Retain only positive activations, enhancing class-relevant pixels.
\end{itemize}
\textbf{Mathematical Formulation} \\
The guided backpropagation gradient is computed as:
\begin{equation}
R^l = \max(0, R^{l+1}) \cdot \max(0, \frac{\partial y}{\partial x})
\end{equation}
where:
\begin{itemize}
    \item \(R^l\)is the relevance at layer l.
\end{itemize}
Guided Backpropagation produces high-resolution visualizations by focusing on fine-grained details. Guided Backpropagation refines standard backpropagation by suppressing negative gradients, ensuring that only features with a positive influence on the output are visualized. This method helps to retain high-resolution details of the input image while focusing on the most relevant pixels.
\section{\textbf{Comparative Analysis of XAI Methods}}
To assess the effectiveness of different XAI methods, I conducted a structured experiment using ResNet50 as the model and applied four interpretability techniques—LIME, SHAP, Grad-CAM, and Guided Backpropagation—on seven diverse species. The selection of the model and species was driven by their interpretability potential across a diverse range of textures, shapes, and patterns.
\subsection*{\textbf{Why ResNet50?}}
For this study, I chose ResNet50 due to its robust architecture and widespread adoption in image classification tasks. The model’s deep residual connections enhance gradient flow, ensuring stable training while enabling effective feature extraction from complex visual inputs. This makes ResNet50 an ideal candidate to evaluate different XAI methods in terms of how they interpret hierarchical feature representations.
\subsection*{\textbf{Why These Seven Species?}}
The selection of these seven species—Samoyed, Maltese, American Robin, Goose, Coyote, Egyptian Cat, and Ladybug—was based on their diversity in textures, shapes, and color patterns. This variety allows a comprehensive evaluation of XAI methods across multiple image types, ranging from fur-covered mammals to feathered birds and insects with distinct body patterns. By incorporating species with varying structural complexities, I aimed to assess how well each XAI method adapts to different visual attributes.
\subsection*{\textbf{Why These Four XAI Methods?}}
I selected LIME and SHAP as they represent model-agnostic approaches, providing interpretability that is independent of the underlying neural architecture. Conversely, Grad-CAM and Guided Backpropagation were chosen as they are model-specific, offering insights into deep-layer feature activations. The combination of these techniques ensures a balanced comparison between local vs. global interpretability, pixel-space vs. feature-space attribution, and perturbation-based vs. gradient-based methods. This selection ensures a balanced evaluation of interpretability trade-offs.
\subsection*{\textbf{Image Grid Representation}}
The following image grid showcases how LIME, SHAP, Grad-CAM, and Guided Backpropagation interpret ResNet50’s predictions on different species. Each row represents a species, and each column represents an XAI method. The analysis following this section breaks down key insights from these visualizations.
\begin{itemize}
    \item Each XAI method interprets the same image differently, providing unique insights into how ResNet50 classifies visual data.
    \item LIME and SHAP assign importance broadly, while Grad-CAM and Guided Backpropagation focus on feature-specific activations.
\end{itemize}

\clearpage
\begin{figure}[htbp]
    \centering
    \renewcommand{\arraystretch}{1.5} 
    \begin{tabular}{ccccc}
        \textbf{Original Image} & \textbf{LIME} & \textbf{SHAP} & \textbf{Grad-CAM} & \textbf{Guided Backpropagation} \\

        \includegraphics[width=3cm, height=3cm, keepaspectratio]{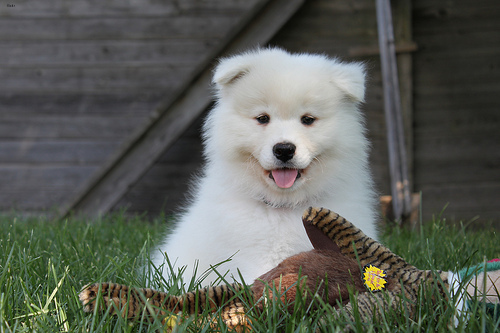} &
        \includegraphics[width=3cm, height=3cm, keepaspectratio]{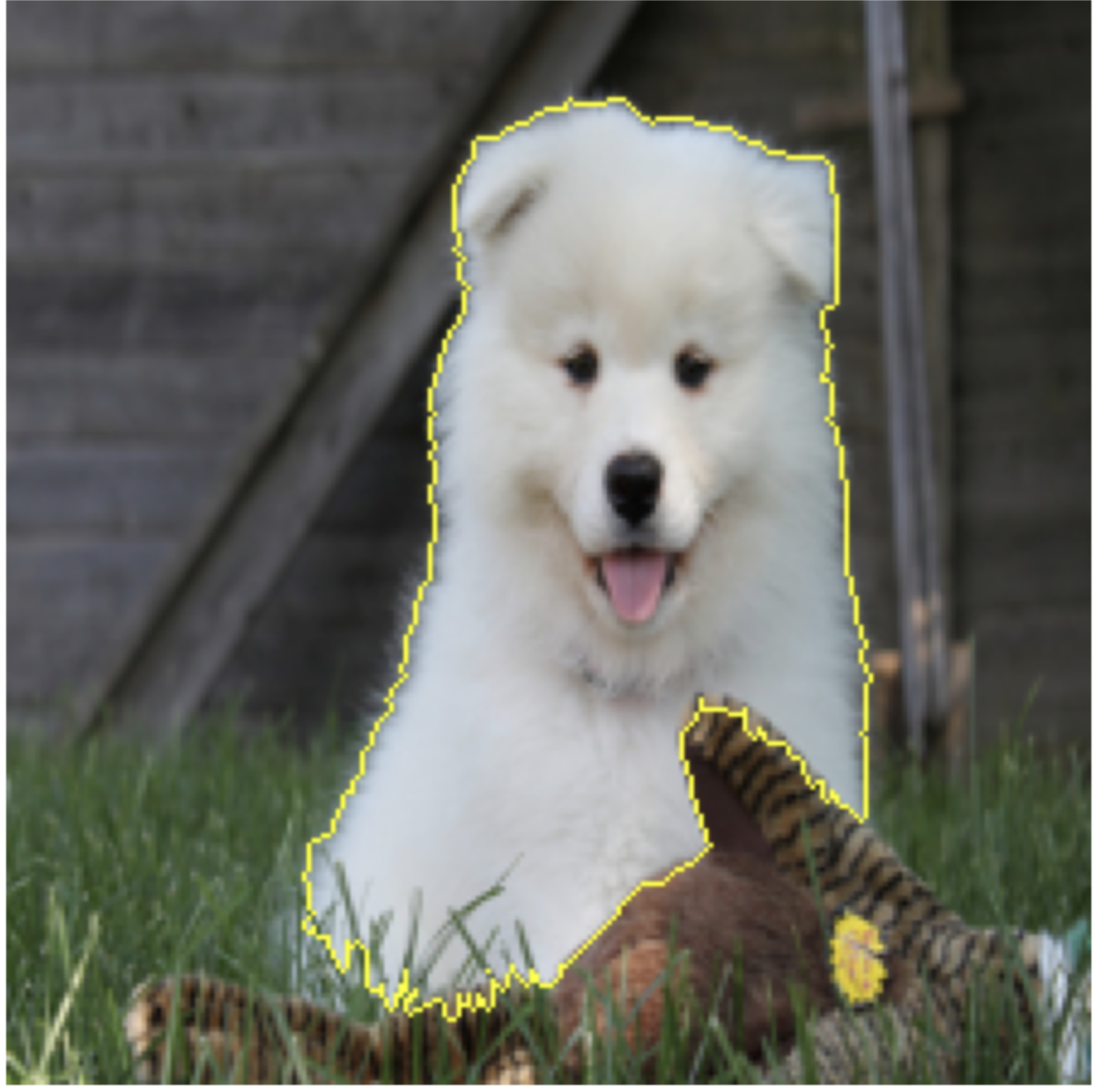} &
        \includegraphics[width=3cm, height=3cm, keepaspectratio]{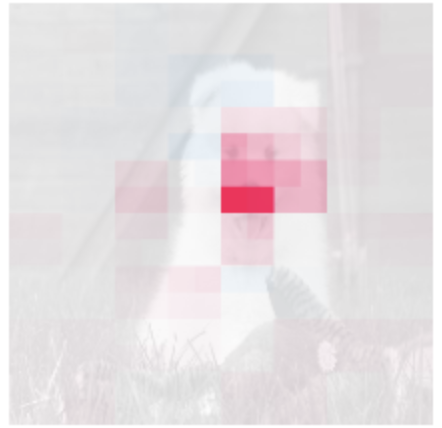} &
        \includegraphics[width=3cm, height=3cm, keepaspectratio]{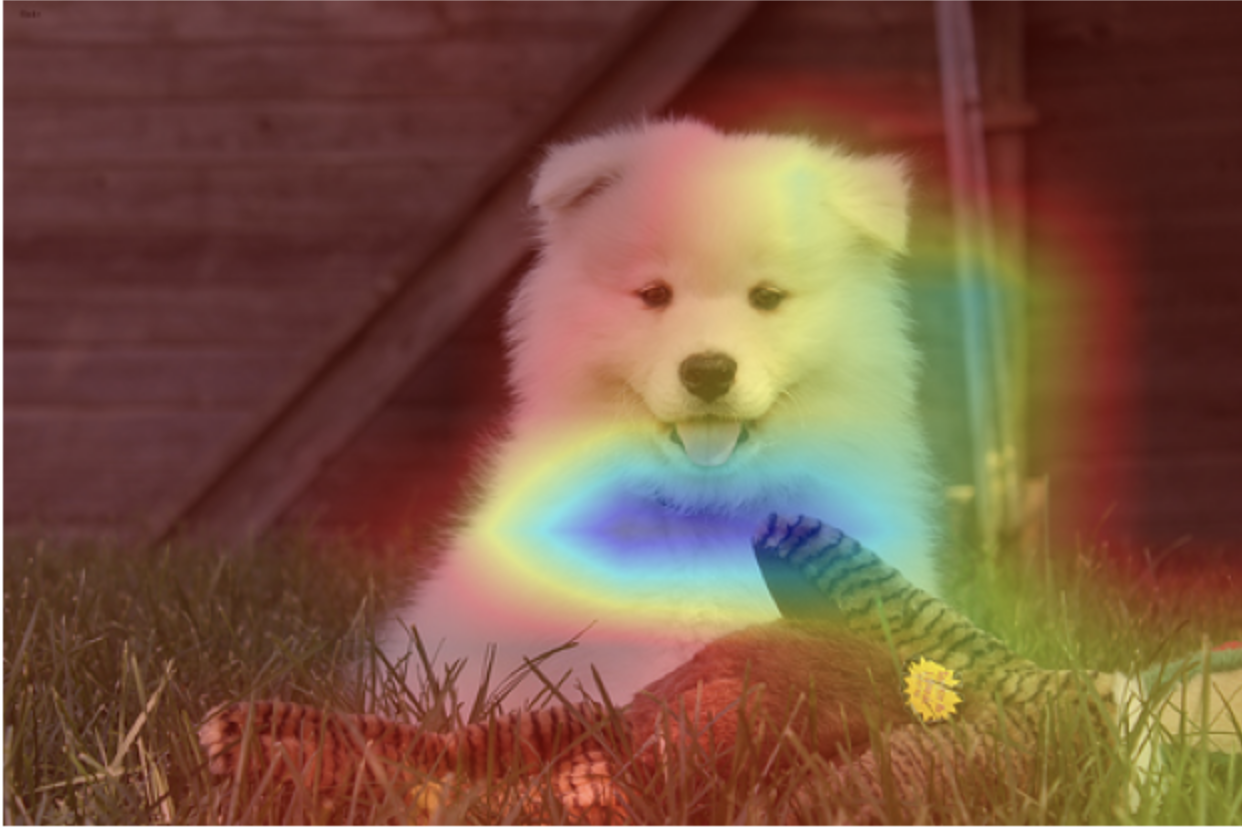} &
        \includegraphics[width=3cm, height=3cm, keepaspectratio]{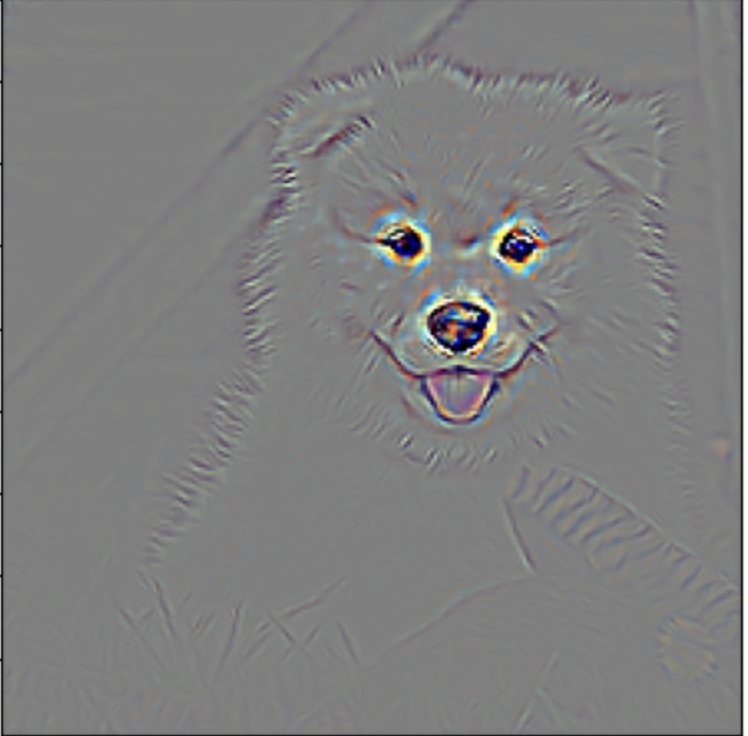} \\

        \includegraphics[width=3cm, height=3cm, keepaspectratio]{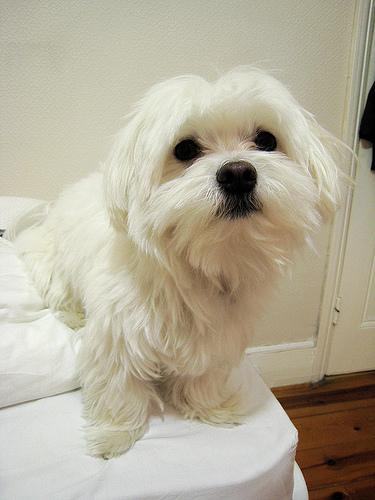} &
        \includegraphics[width=3cm, height=3cm, keepaspectratio]{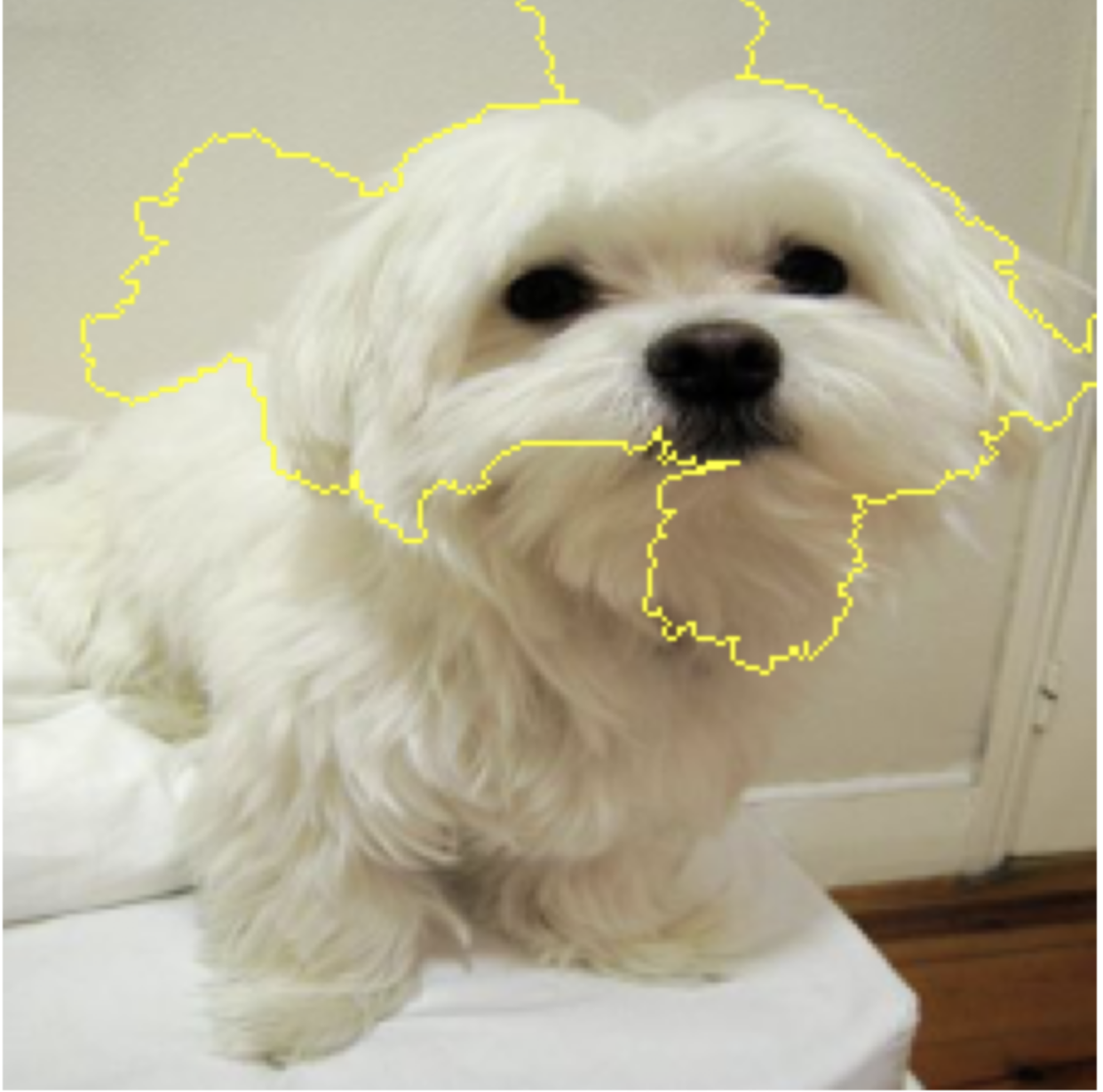} &
        \includegraphics[width=3cm, height=3cm, keepaspectratio]{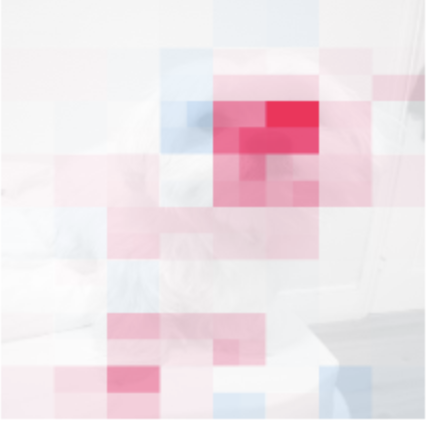} &
        \includegraphics[width=3cm, height=3cm, keepaspectratio]{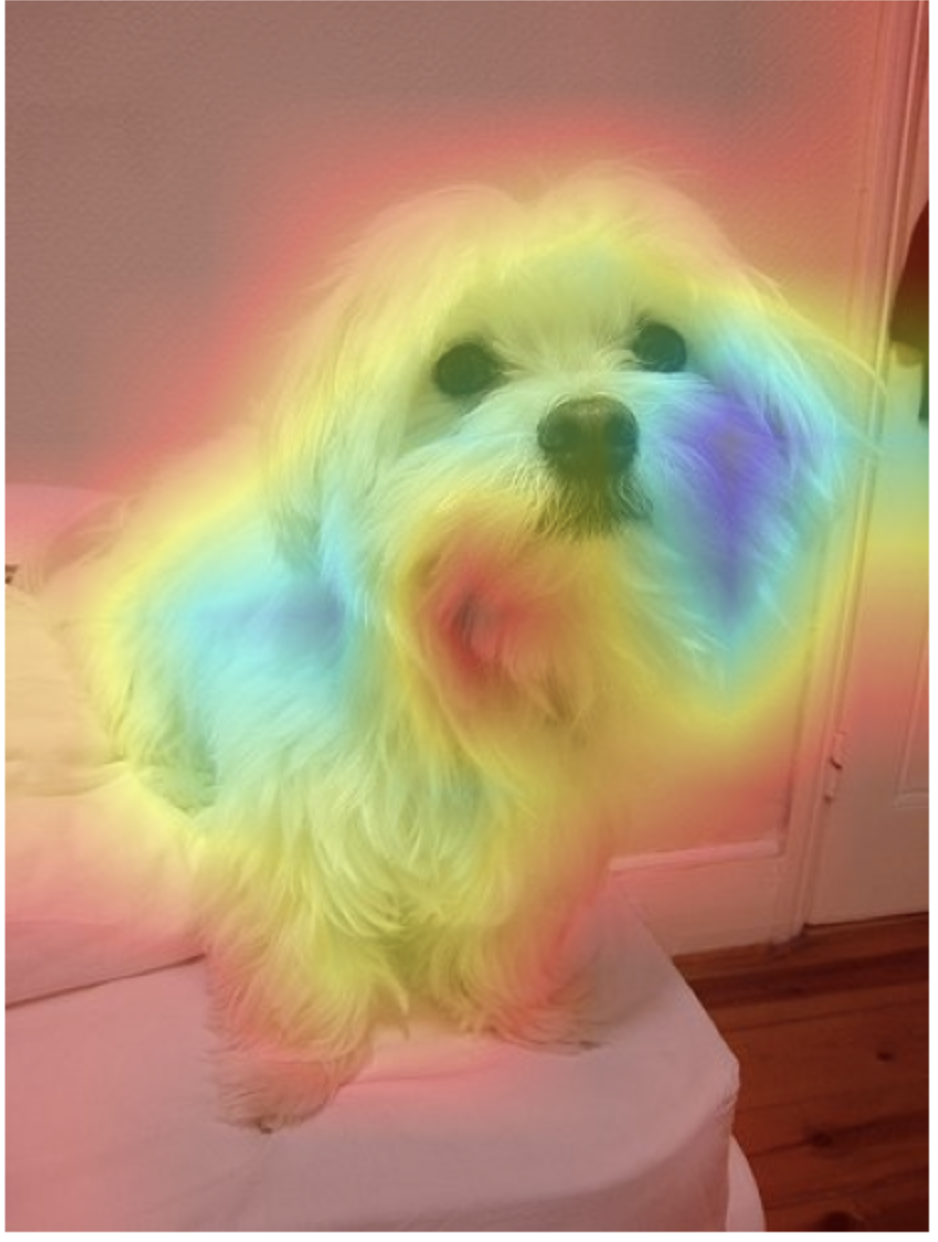} &
        \includegraphics[width=3cm, height=3cm, keepaspectratio]{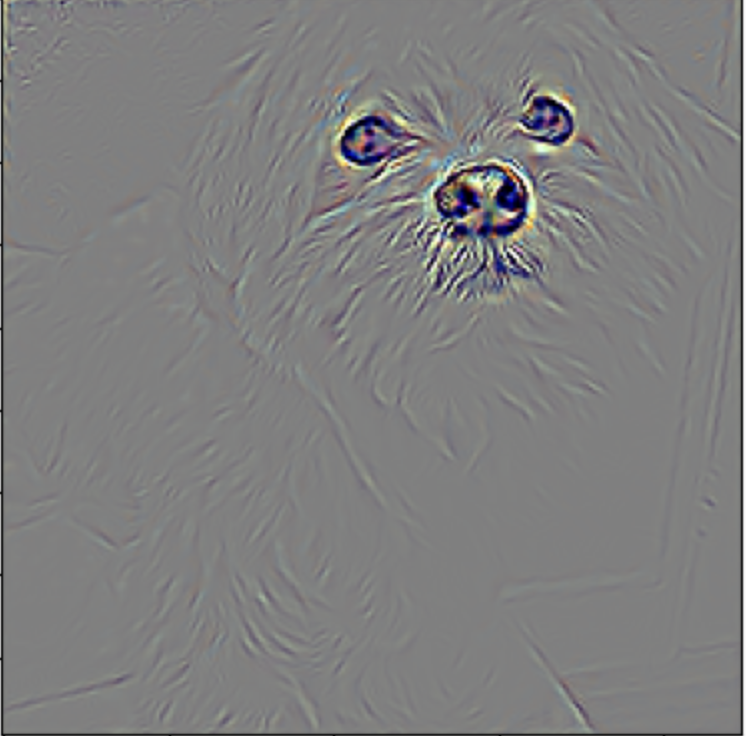} \\

        \includegraphics[width=3cm, height=3cm, keepaspectratio]{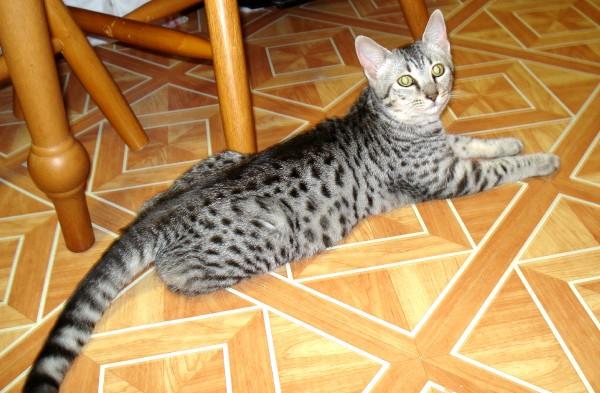} &
        \includegraphics[width=3cm, height=3cm, keepaspectratio]{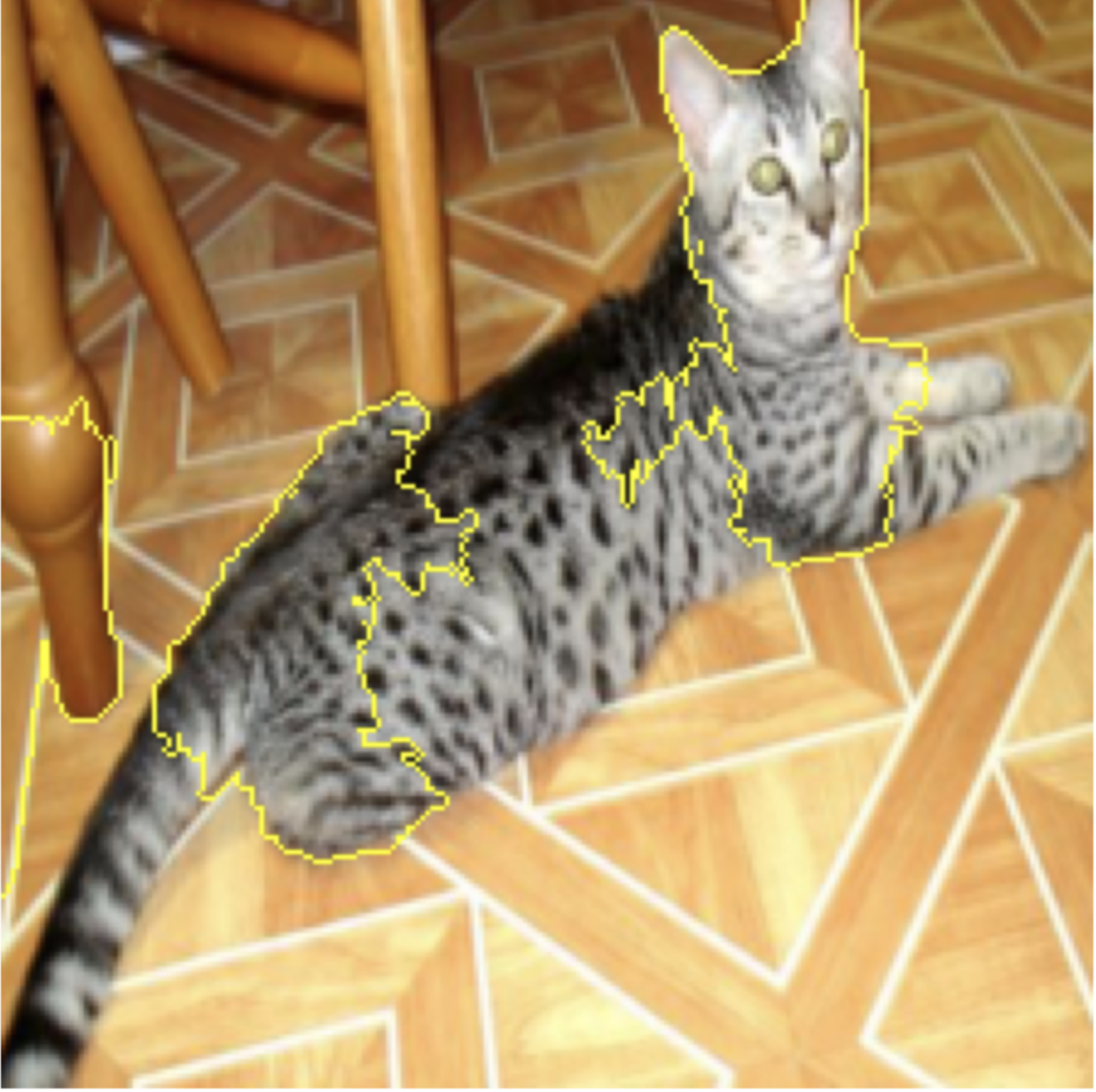} &
        \includegraphics[width=3cm, height=3cm, keepaspectratio]{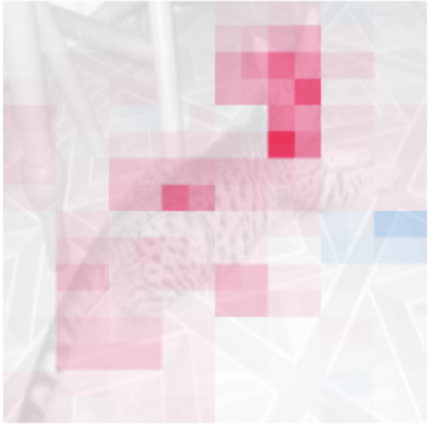} &
        \includegraphics[width=3cm, height=3cm, keepaspectratio]{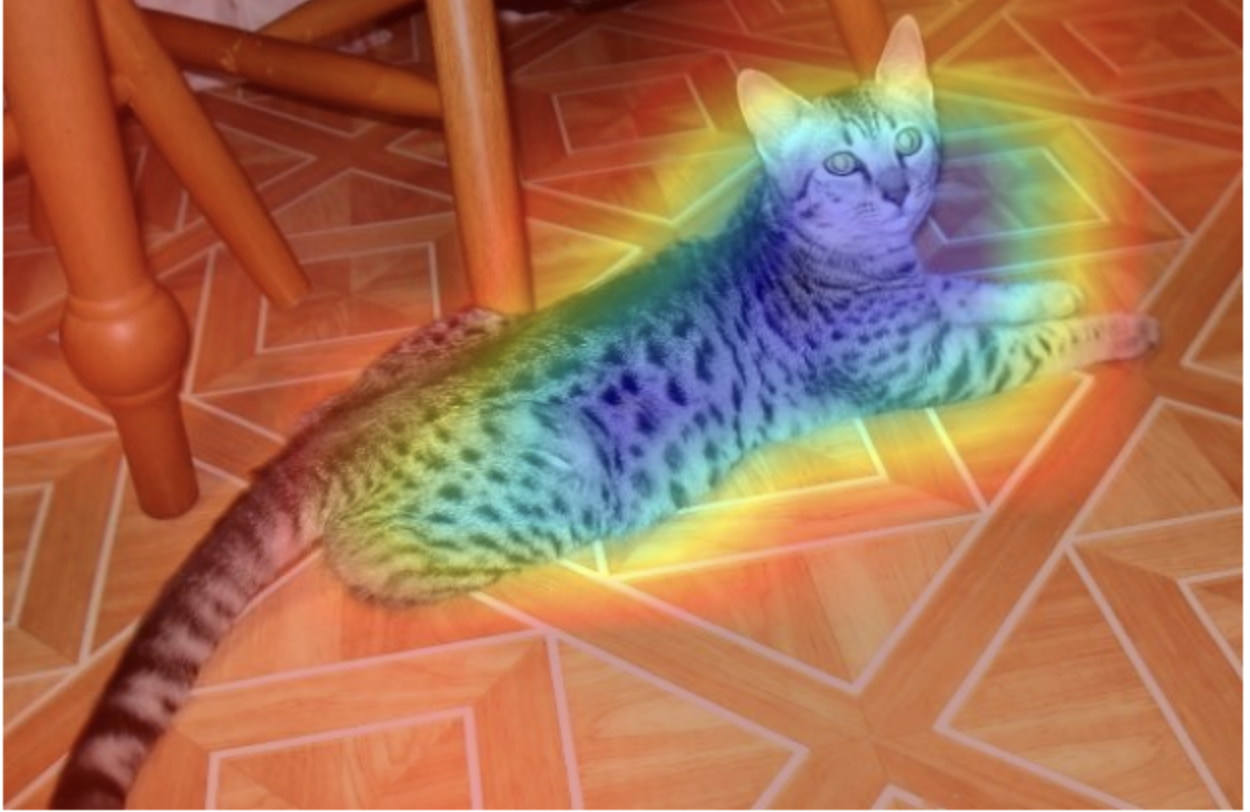} &
        \includegraphics[width=3cm, height=3cm, keepaspectratio]{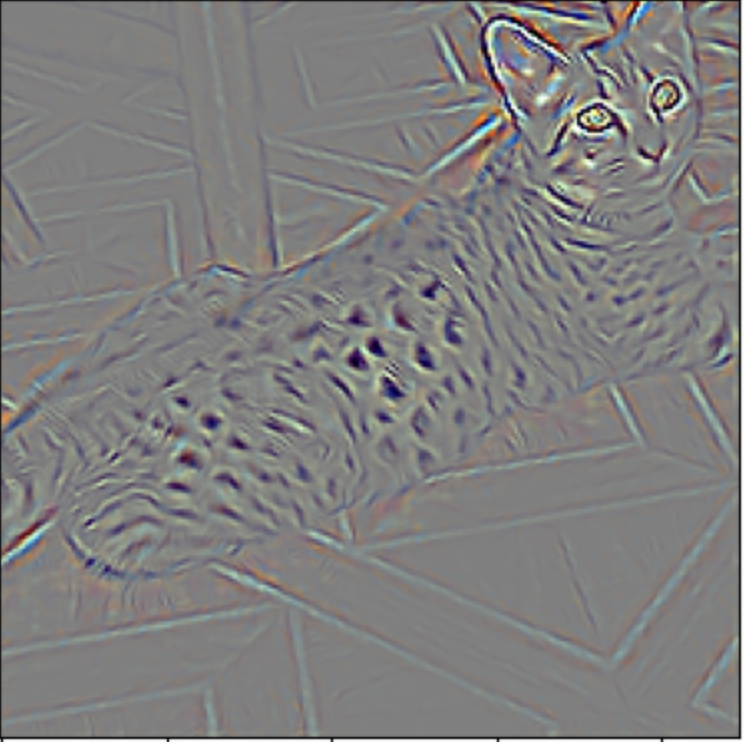} \\

        \includegraphics[width=3cm, height=3cm, keepaspectratio]{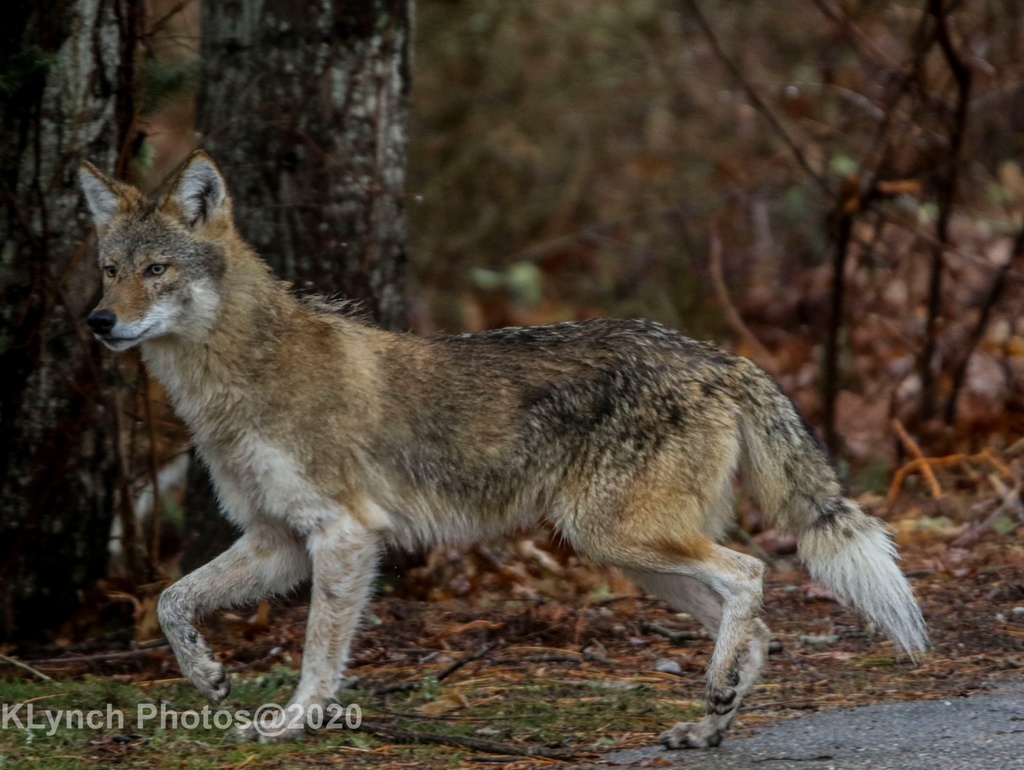} &
        \includegraphics[width=3cm, height=3cm, keepaspectratio]{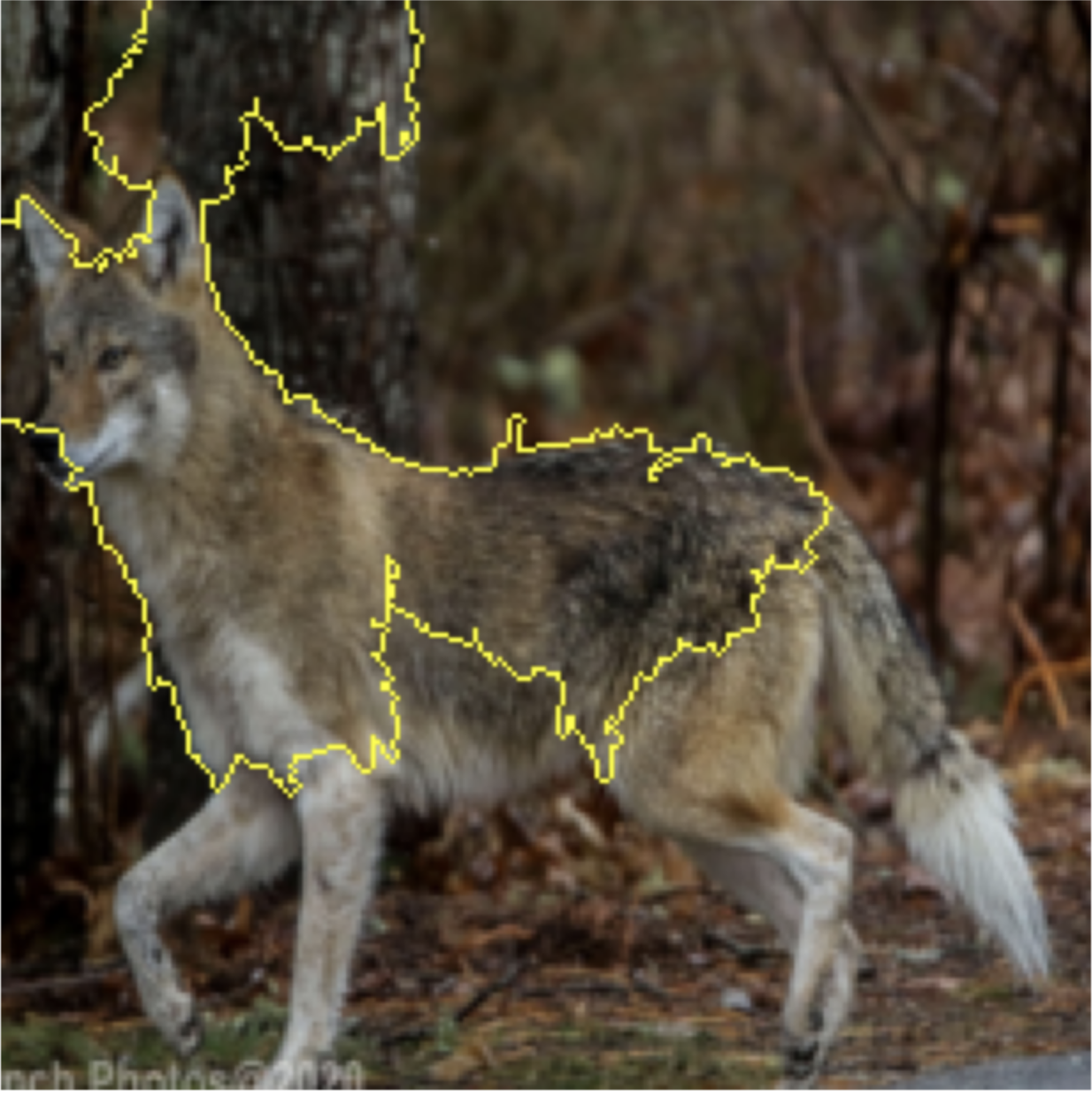} &
        \includegraphics[width=3cm, height=3cm, keepaspectratio]{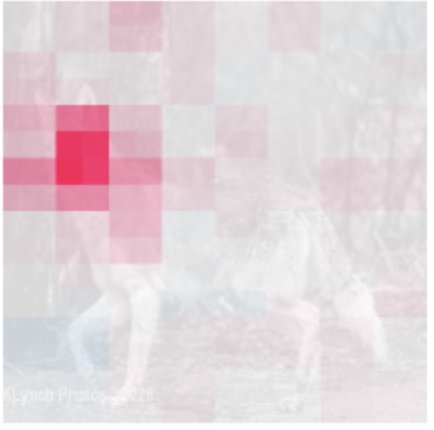} &
        \includegraphics[width=3cm, height=3cm, keepaspectratio]{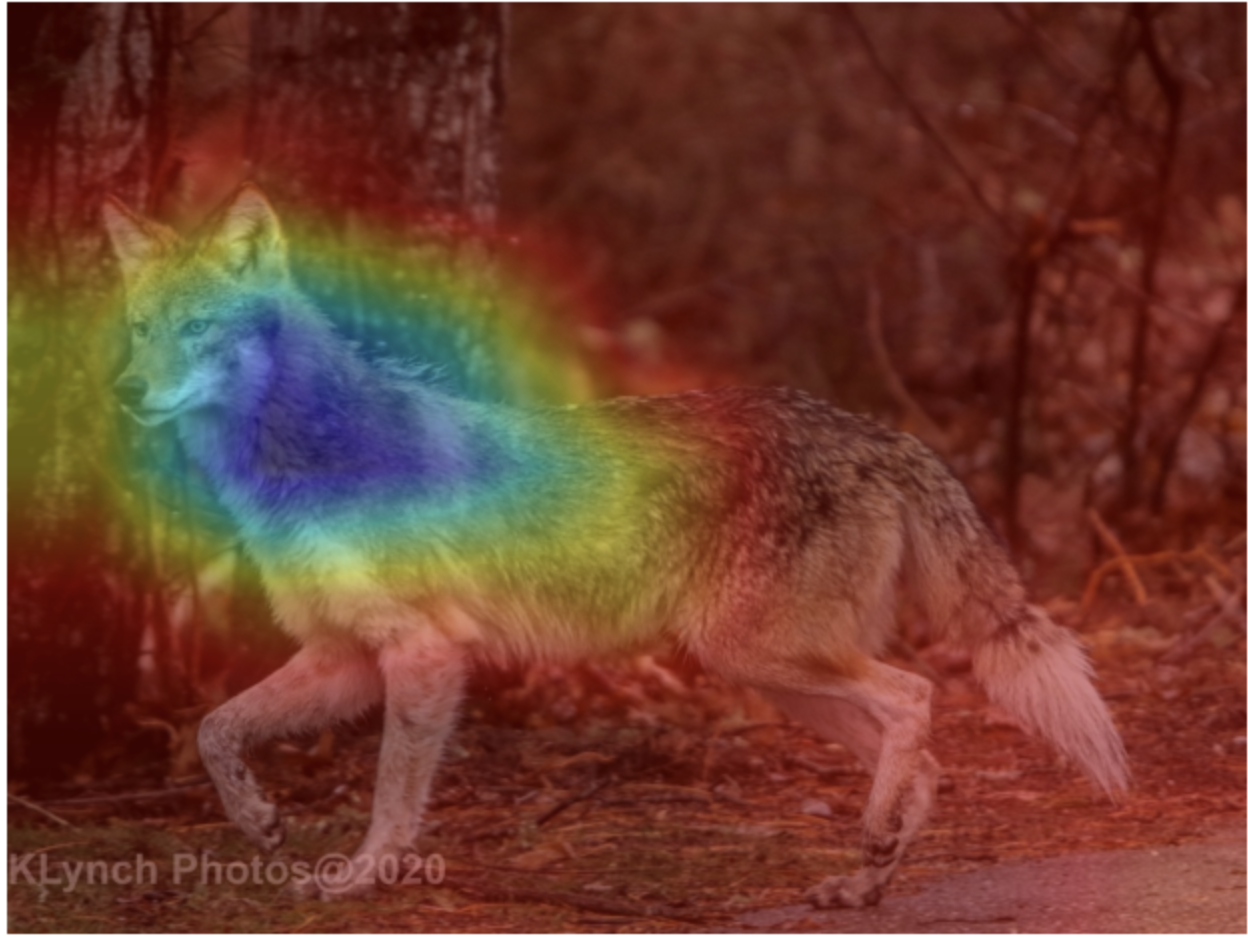} &
        \includegraphics[width=3cm, height=3cm, keepaspectratio]{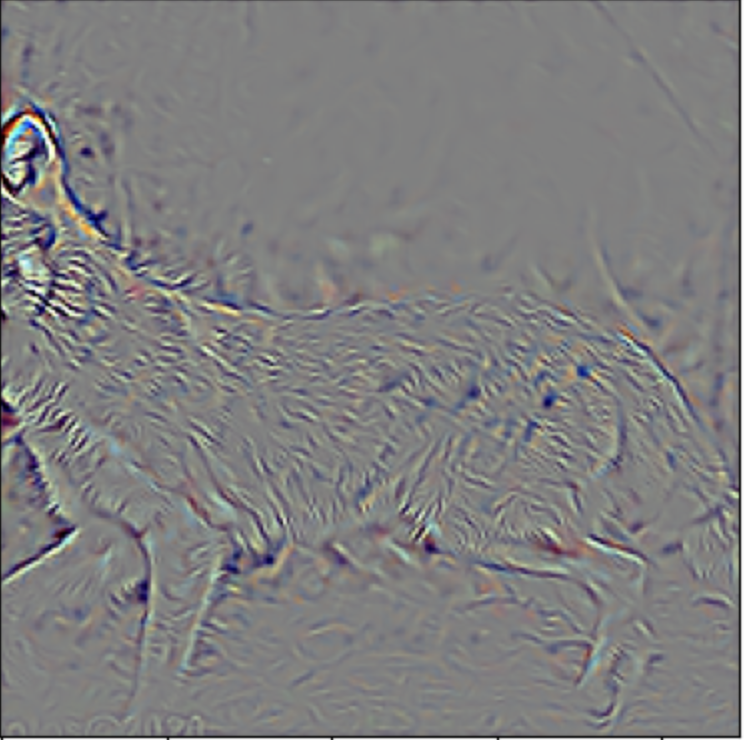} \\

        \includegraphics[width=3cm, height=3cm, keepaspectratio]{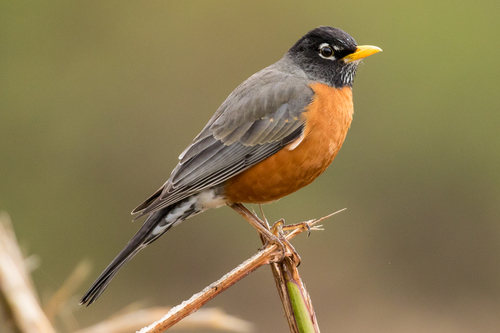} &
        \includegraphics[width=3cm, height=3cm, keepaspectratio]{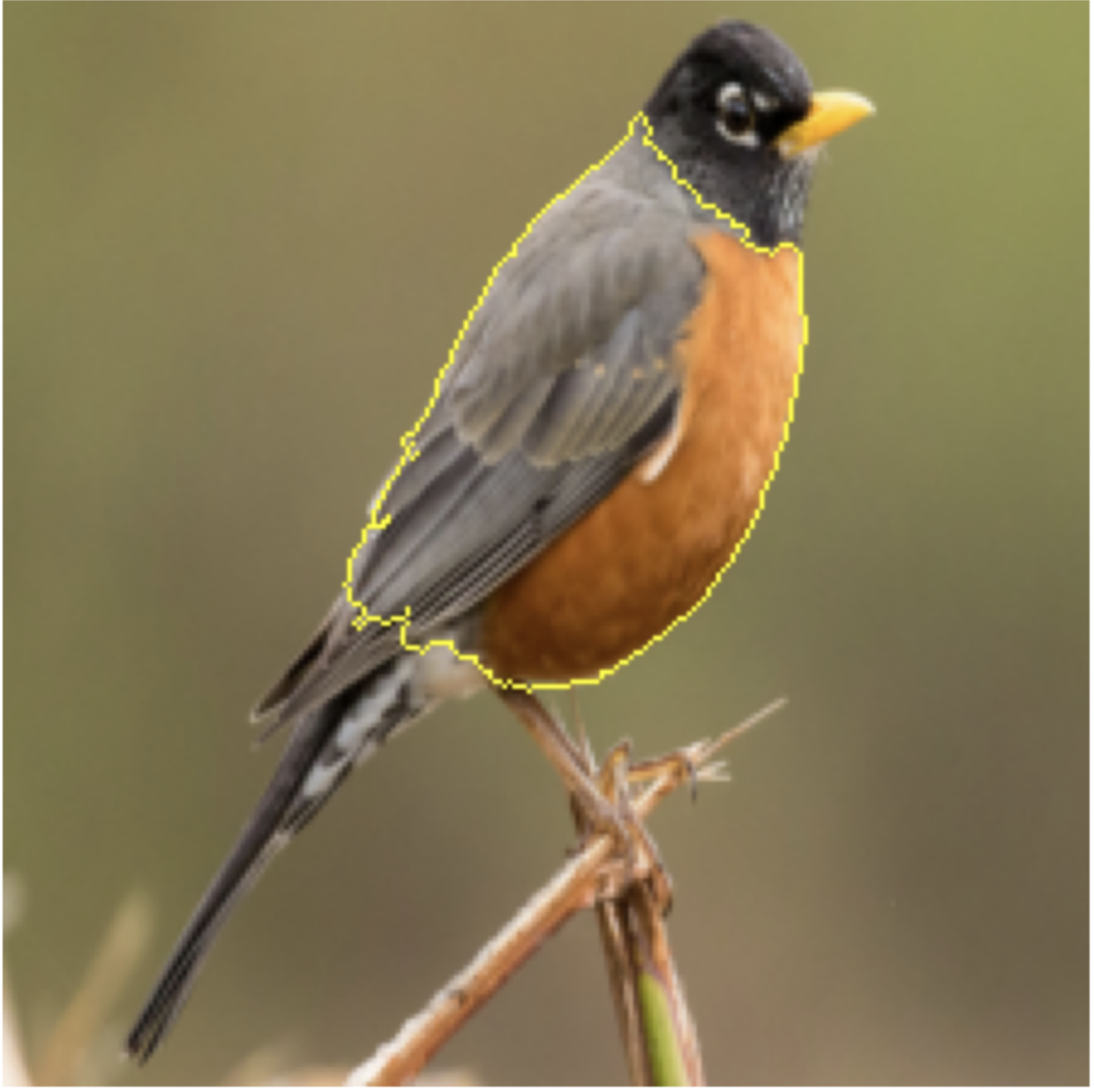} &
        \includegraphics[width=3cm, height=3cm, keepaspectratio]{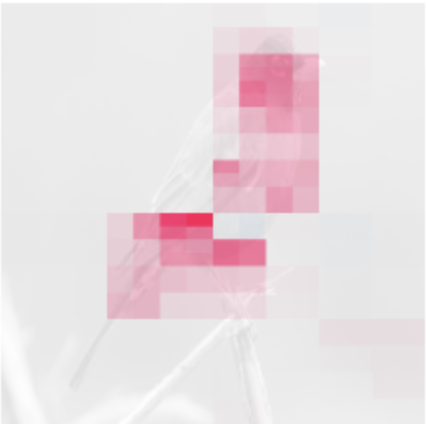} &
        \includegraphics[width=3cm, height=3cm, keepaspectratio]{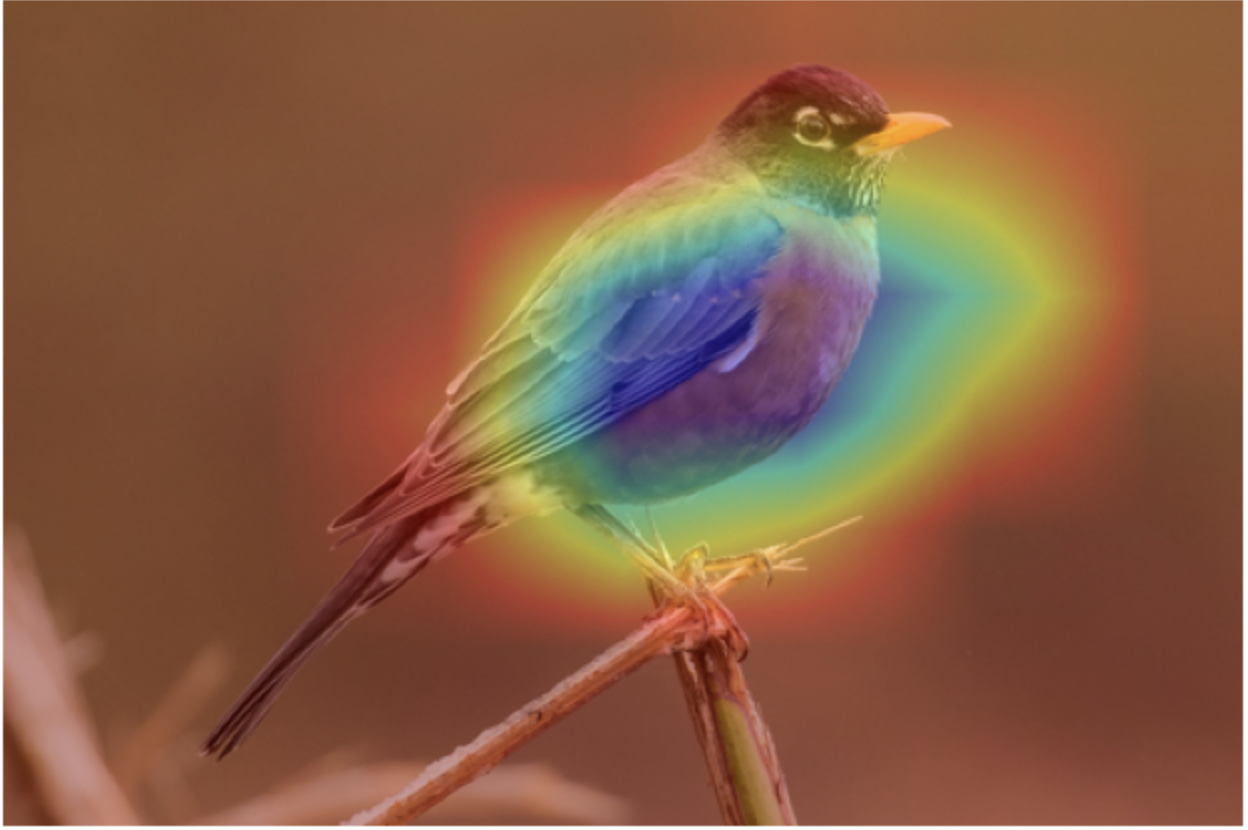} &
        \includegraphics[width=3cm, height=3cm, keepaspectratio]{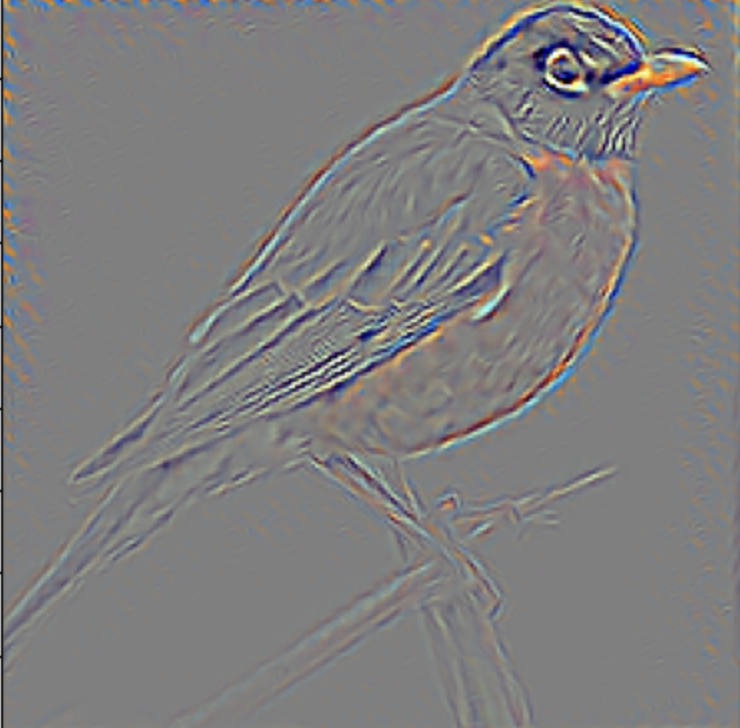} \\

        \includegraphics[width=3cm, height=3cm, keepaspectratio]{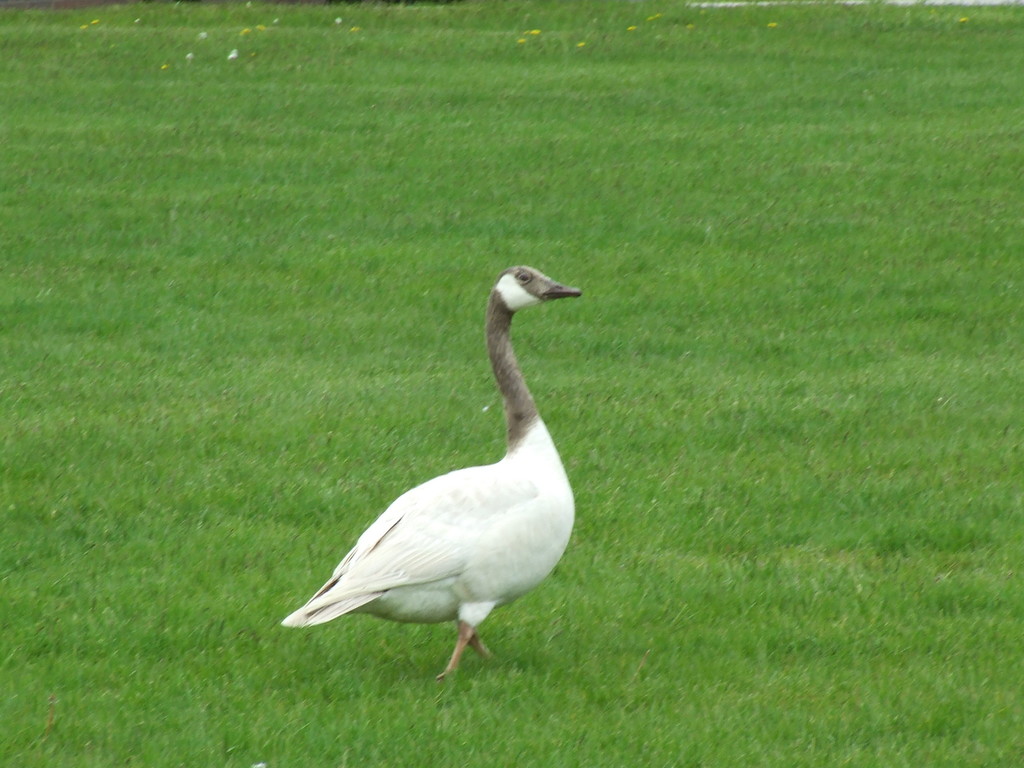} &
        \includegraphics[width=3cm, height=3cm, keepaspectratio]{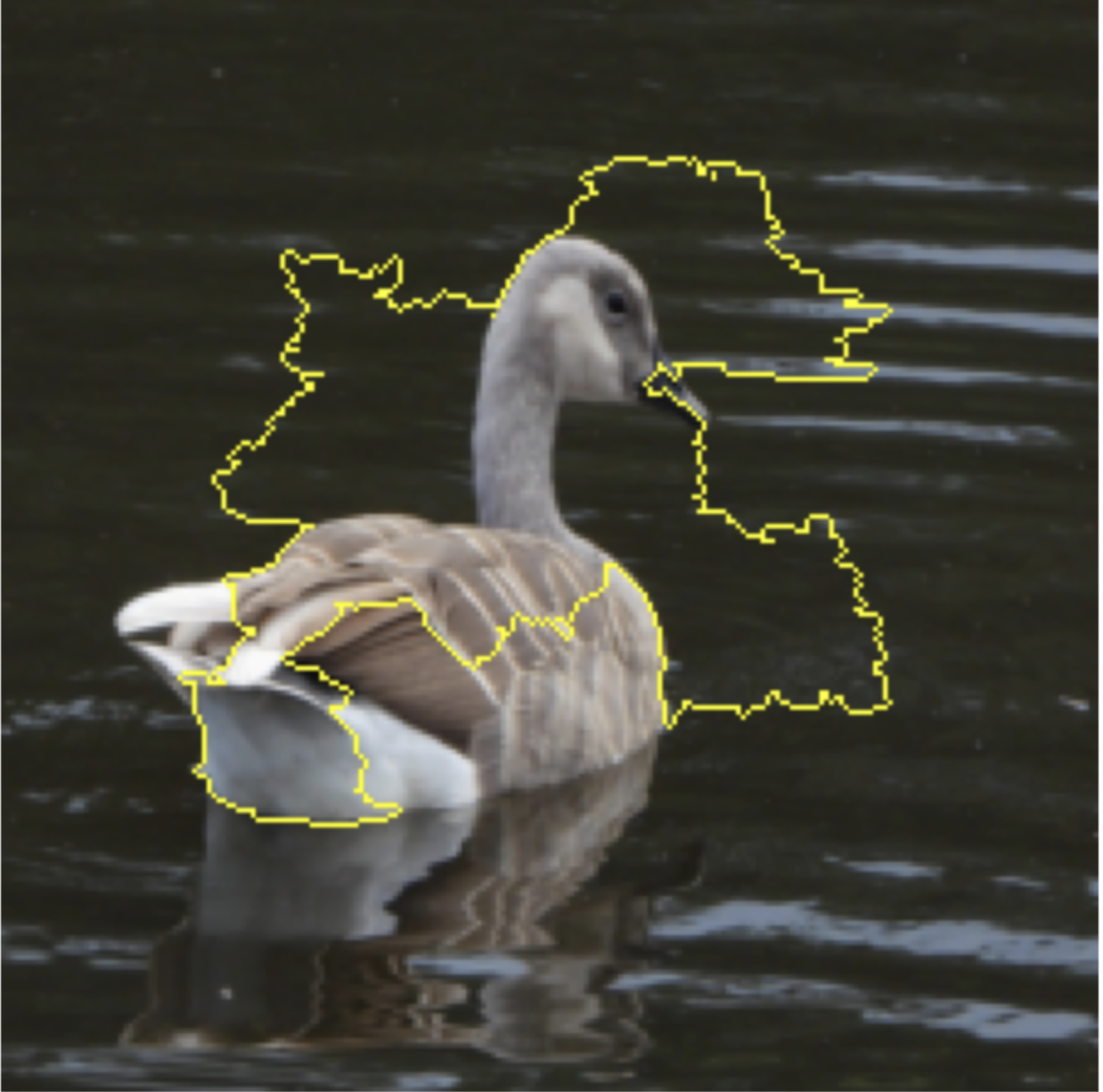} &
        \includegraphics[width=3cm, height=3cm, keepaspectratio]{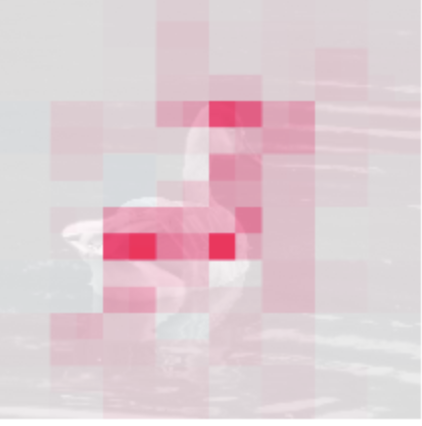} &
        \includegraphics[width=3cm, height=3cm, keepaspectratio]{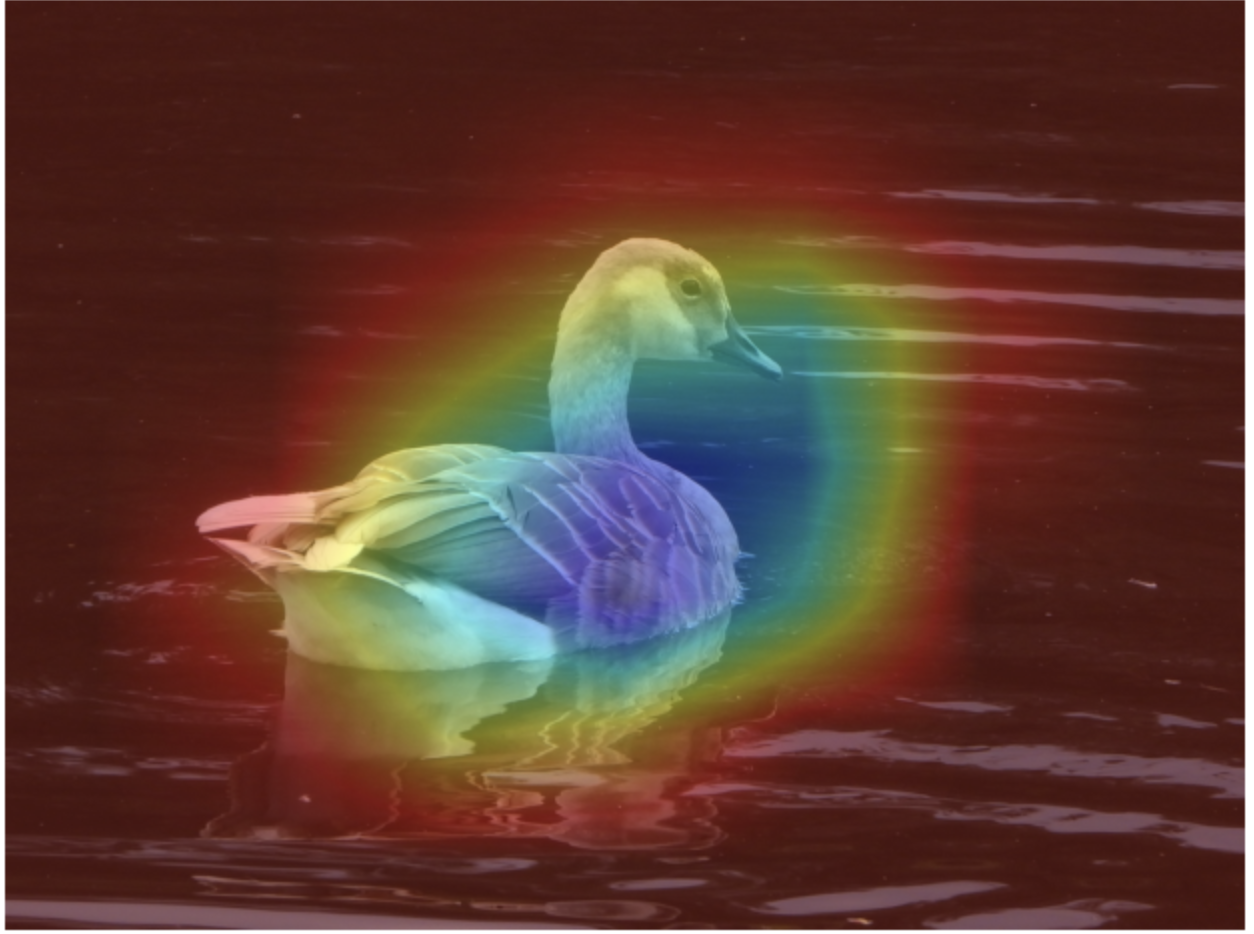} &
        \includegraphics[width=3cm, height=3cm, keepaspectratio]{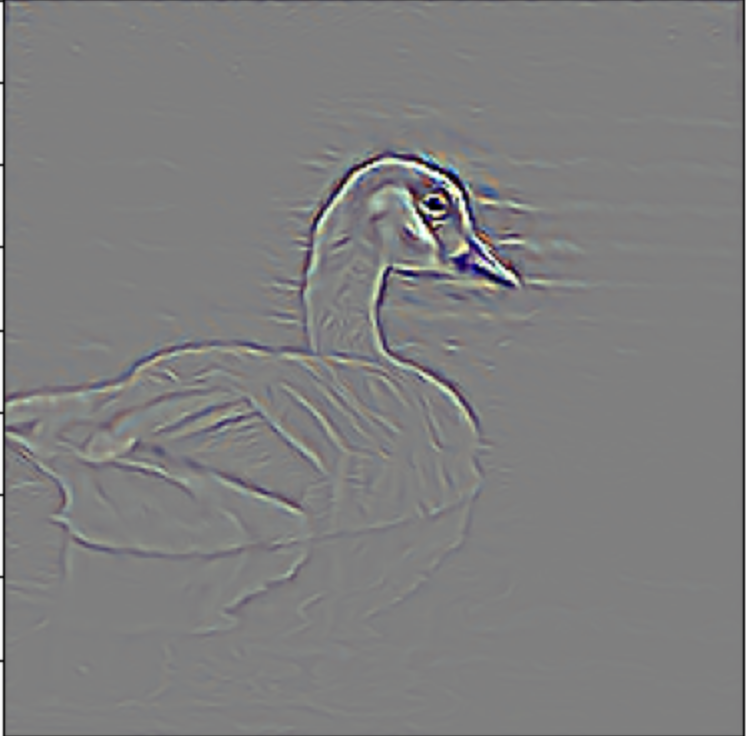} \\

        \includegraphics[width=3cm, height=3cm, keepaspectratio]{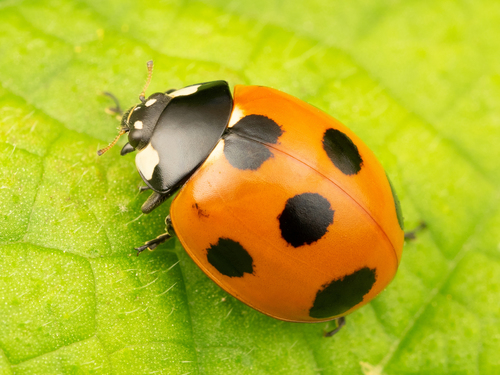} &
        \includegraphics[width=3cm, height=3cm, keepaspectratio]{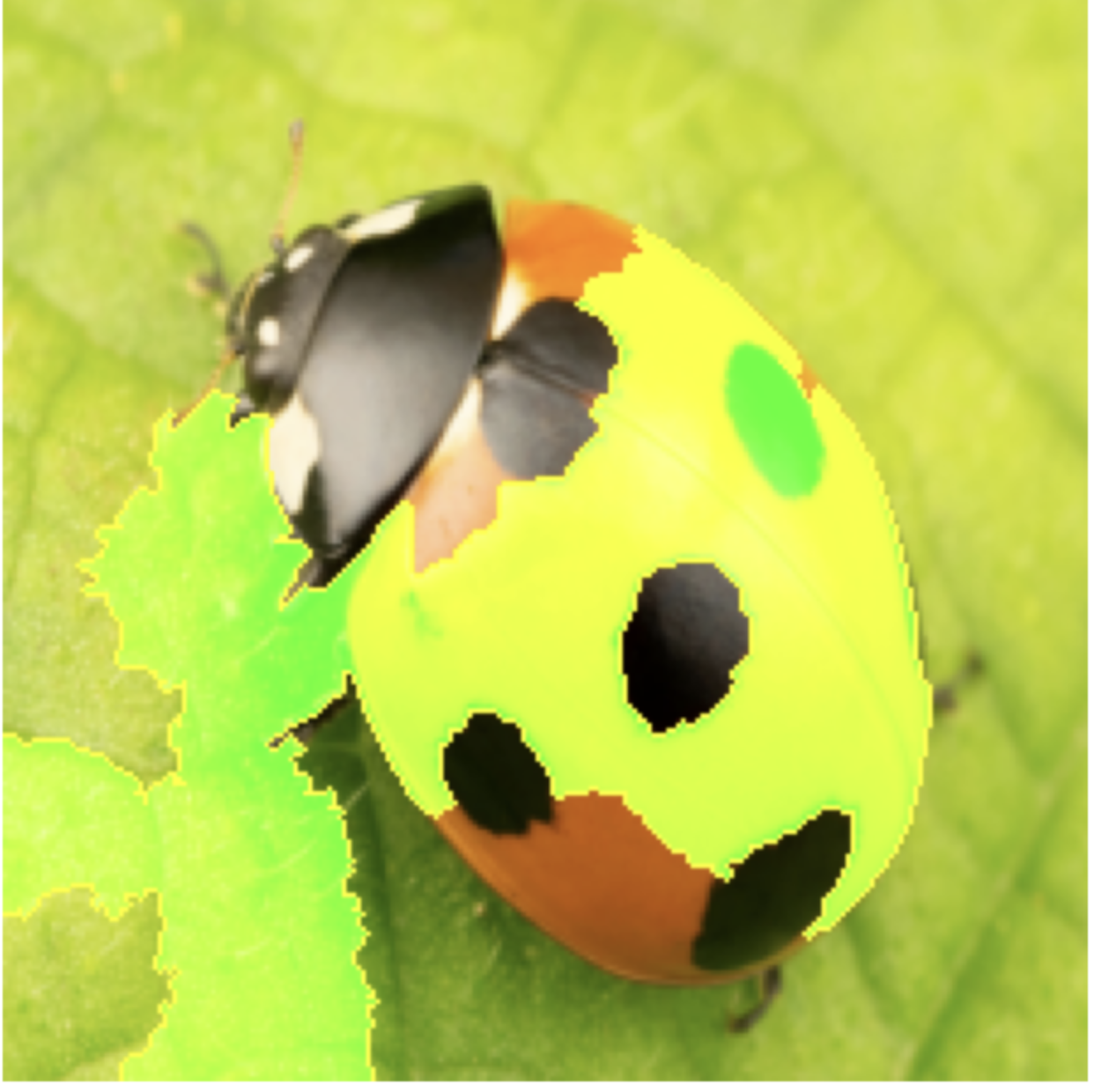} &
        \includegraphics[width=3cm, height=3cm, keepaspectratio]{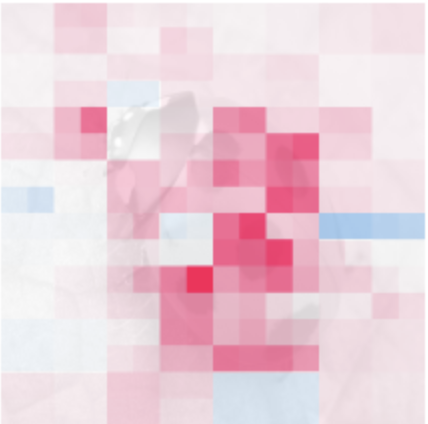} &
        \includegraphics[width=3cm, height=3cm, keepaspectratio]{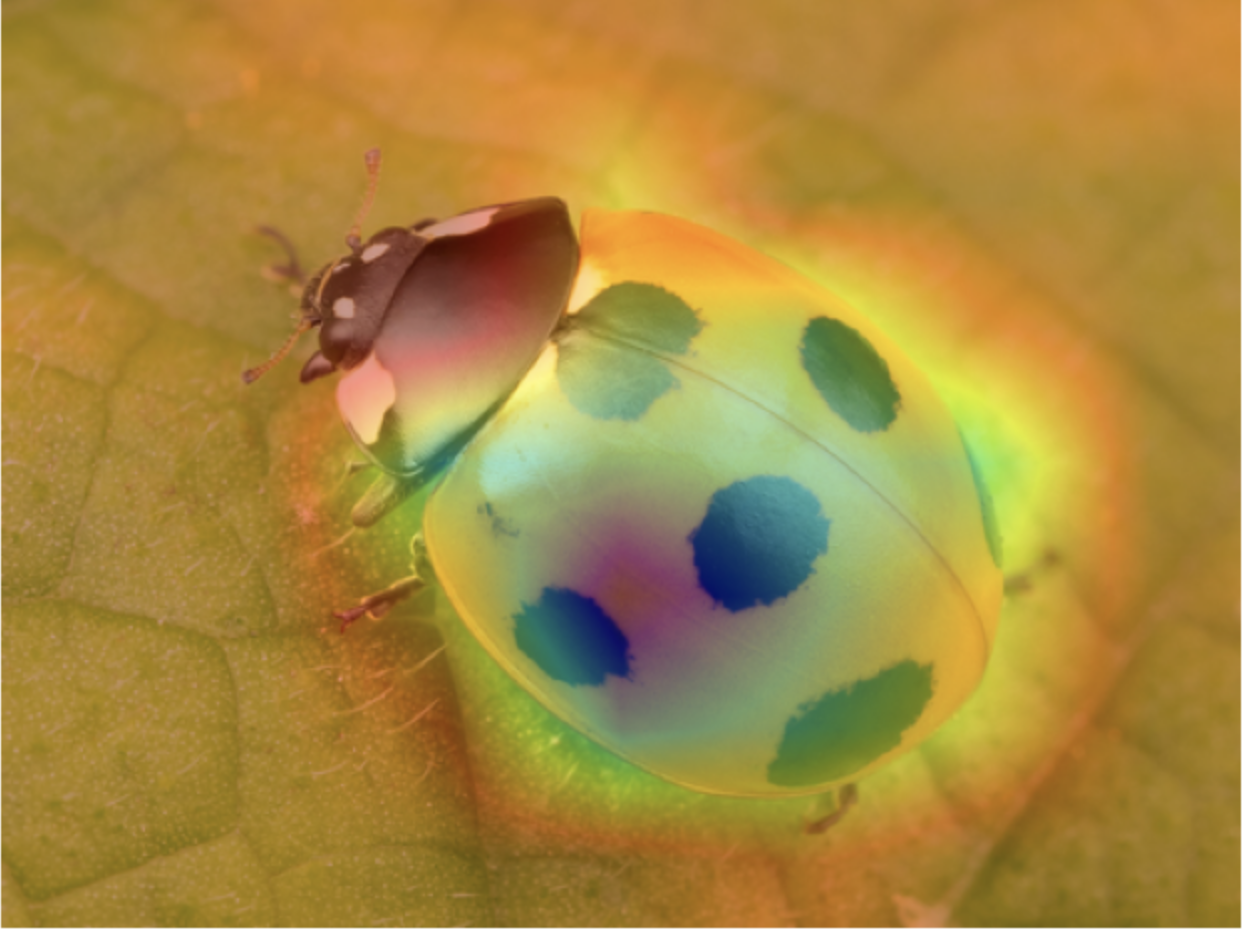} &
        \includegraphics[width=3cm, height=3cm, keepaspectratio]{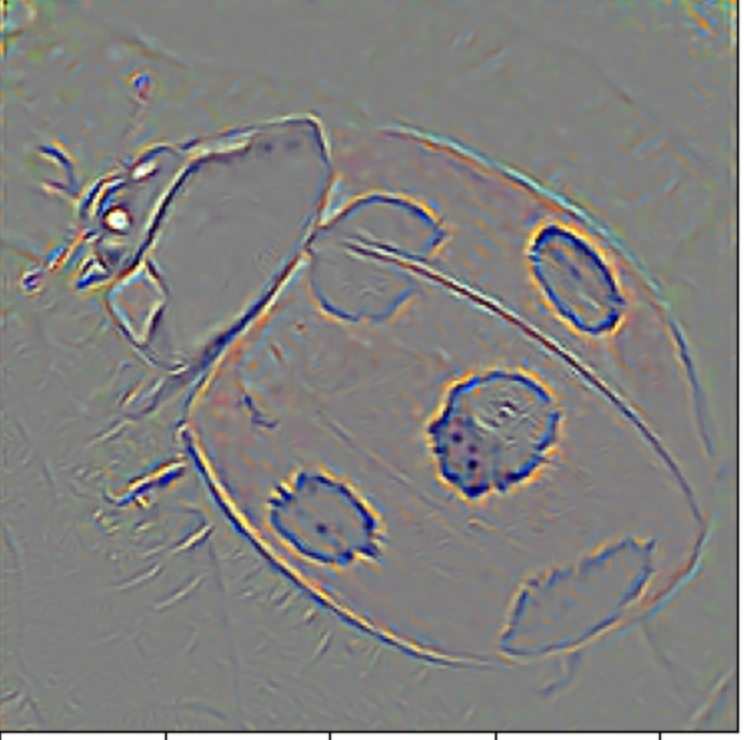} \\

    \end{tabular}
    \caption{Comparative visualizations of XAI methods applied to different species. Each row represents a species, and each column represents a different XAI method.}
    \label{fig:xai_grid}
\end{figure}
\clearpage
\subsection*{\textbf{Key Observations and Comparative Insights}}
Here is what I observed on performing robust analysis, as mentioned tested these techniques on diverse categories, including dogs (Samoyed, Maltese), birds (American Robin, Goose), wild animals (Coyote, Egyptian Cat), and insects (Ladybug).
\begin{table}[h]
    \centering
    \begin{tabular}{|p{2cm}|p{2cm}|p{4cm}|p{4cm}|p{4cm}|}
    \hline
    \textbf{Method} & \textbf{Type} & \textbf{Strengths} & \textbf{Weaknesses} & \textbf{Best Used For} \\ \hline
    LIME & Model-Agnostic & Works with any model, provides interpretable explanations & Sensitive to perturbation size, explanations vary across runs & Feature importance for any ML model \\ \hline
    SHAP & Model-Agnostic & Theoretically sound, consistent global feature attribution & Computationally expensive, slow for deep models & Understanding feature contributions in high-dimensional data \\ \hline
    Grad-CAM & Model-Specific (CNNs) & Localizes key discriminative regions, preserves spatial importance & Only applicable to convolutional layers, lacks fine-grained details & Image-based deep learning tasks \\ \hline
    Guided Backpropagation & Model-Specific (CNNs) & Produces high-resolution feature visualizations & Does not highlight class-specific importance, can be noisy & Detailed edge and texture analysis \\ \hline
    \end{tabular}
    \label{tab:explanation_methods}
\end{table}
\subsection*{\textbf{Comparing Across the Seven Images}}
\begin{enumerate}
    \item \textbf{Dogs (Samoyed, Maltese):}
\end{enumerate}
\begin{itemize}
    \item LIME \& SHAP clearly highlighted the entire body contour, reinforcing the importance of texture and shape in classification.
    \item Grad-CAM, however, tended to focus on the face, particularly the eyes and nose, suggesting that the model prioritizes these features.
    \item Guided Backpropagation emphasized fine-grained textures, making the fur patterns highly visible.
\end{itemize}
\begin{enumerate}
    \setcounter{enumi}{1}
    \item \textbf{Wild Animals (Coyote, Egyptian Cat):}
\end{enumerate}
\begin{itemize}
    \item SHAP and LIME highlighted the overall body of the animals but also included background elements, indicating potential biases in the model.
    \item Grad-CAM remained class-discriminative, capturing the torso and facial regions, reinforcing how the CNN processes structural patterns.
    \item Guided Backpropagation excelled in edge detection, clearly outlining the fur textures and body contours.
\end{itemize}
\begin{enumerate}
    \setcounter{enumi}{2}
    \item \textbf{Birds (American Robin, Goose):}
\end{enumerate}
\begin{itemize}
    \item Grad-CAM showed a strong focus on the head and beak, confirming that these areas contribute heavily to the classification.
    \item SHAP and LIME distributed importance more broadly across the entire body, making them better at holistic feature attribution.
    \item Guided Backpropagation distinctly highlighted the feather patterns, showcasing the fine-grained texture dependency in CNN decision-making.
\end{itemize}

\begin{enumerate}
    \setcounter{enumi}{3}
    \item \textbf{Insect (Ladybug):}
\end{enumerate}
\begin{itemize}
    \item LIME \& SHAP primarily highlighted the red shell with black spots, confirming that color and pattern are crucial.
    \item Grad-CAM focused on the upper half, suggesting antennae and head structures were primary cues.
    \item Guided Backpropagation precisely outlined the shell edges, confirming that the model recognizes the overall contour distinctly.
\end{itemize}
\subsection*{\textbf{Trade-Offs Between Model-Agnostic \& Model-Specific Methods}}
\subsubsection*{\textbf{Generalization vs. Model Constraints}}
\begin{itemize}
    \item Model-agnostic methods (LIME, SHAP) provide generalizable explanations but can be computationally expensive.
    \item Model-specific methods (Grad-CAM, Guided Backpropagation) are efficient but constrained to certain architectures.
\end{itemize}
\subsubsection*{\textbf{Feature Attribution Differences}}
\begin{itemize}
    \item SHAP consistently provides global importance, whereas LIME focuses on localized perturbation-based explanations.
    \item Grad-CAM highlights activation-heavy regions, while Guided Backpropagation enhances edge detection and texture focus.
\end{itemize}

\subsubsection*{\textbf{Computational Efficiency Matters}}
\begin{itemize}
    \item SHAP is computationally intensive but highly reliable.
    \item Grad-CAM and Guided Backpropagation provide faster interpretations suited for real-time scenarios.
\end{itemize}

\subsection*{\textbf{Final Reflections on XAI Methodology}}
This comparative analysis demonstrates that no single XAI method is universally superior. Instead, their effectiveness depends on the problem, model architecture, and computational constraints:

\begin{itemize}
    \item LIME \& SHAP - Broad feature attribution across different models.
    \item Grad-CAM \& Guided Backpropagation - Deep learning-focused with spatial activation insights.
    \item Hybrid Approaches (Combining multiple methods) provide the most robust explainability.
\end{itemize}
This study underscores the importance of multi-faceted XAI methodologies to achieve a comprehensive and reliable interpretation of AI models.
\section{\textbf{Conclusion \& Future Work}}
In this study, I systematically explored and compared different XAI methods—LIME, SHAP, Grad-CAM, and Guided Backpropagation—by applying them to ResNet50 across seven diverse species. The results demonstrated that each method provides unique insights into the model’s decision-making process, with model-agnostic techniques (LIME, SHAP) offering broader feature attribution, while model-specific techniques (Grad-CAM, Guided Backpropagation) emphasized localized, class-relevant areas.
\\
A key takeaway from this comparative analysis is that no single explainability method suffices in all scenarios. Instead, combining multiple XAI techniques provides a more comprehensive understanding of model behavior. For example, using SHAP for global interpretability and Grad-CAM for class-discriminative localization ensures both holistic and fine-grained insights.
Looking ahead, future work can focus on the following directions:
\begin{itemize}
    \item Combining Model-Agnostic and Model-Specific Methods: Hybrid approaches that leverage both perturbation-based and gradient-based techniques can enhance interpretability while mitigating individual method limitations.
    \item Scalability in Large-Scale Datasets: Exploring optimized implementations of computationally expensive methods (such as SHAP) to improve efficiency for real-time AI applications.
    \item Explainability Beyond CNNs: Extending model-specific interpretability techniques to non-convolutional architectures, such as transformers and graph neural networks, to broaden applicability.
    \item Human-Centric Interpretability: Evaluating how well these explanations align with human intuition and domain expertise, ensuring their practical usability in high-stakes applications like healthcare and autonomous systems.
\end{itemize}
As XAI continues to evolve, integrating multiple interpretability approaches while addressing computational efficiency and real-world usability will be crucial for building transparent, reliable, and trustworthy AI systems.

\end{document}